\documentclass[journal]{IEEEtran}
\pdfoutput=1
\usepackage{hyperref}
\usepackage{url}
\usepackage{graphicx}
\usepackage{multirow}
\usepackage{CJK}
\usepackage{bm}
\usepackage{times}
\usepackage{amsmath,amsfonts}
\usepackage[thmmarks,amsmath]{ntheorem}

\qedsymbol{\ensuremath{\square}}
   \theoremstyle{nonumberplain}
   \theoremheaderfont{\bfseries}
   \theorembodyfont{\normalfont}
   \theoremsymbol{\ensuremath{\square}}
   \newtheorem{proof}{Proof}

\usepackage{textcomp}
\usepackage{stfloats}
\usepackage{caption}
\usepackage{cite}

\begin{document}

\title{Stationary Point Losses for Robust Model}

\author{Weiwei Gao,
        Dazhi Zhang,
        Yao Li,
        Zhichang Guo,
        Ovanes Petrosian
\thanks{\S Yao Li is corresponding author. (E-mail: yaoli0508@hit.edu.cn).}
\thanks{Weiwei Gao, Dazhi Zhang,Yao Li and Zhichang Guo are with the Department of Computational Mathematics, School of Mathematics, Harbin Institute of Technology, Harbin 150001, China.}

\thanks{Ovanes Petrosian is with the Department of Mathematical Game Theory and Statistical Decisions, Saint-Petersburg State University, Russian}}
\maketitle

\begin{abstract}
The inability to guarantee robustness is one of the major obstacles to the application of deep learning models in security-demanding domains. We identify that the most commonly used cross-entropy (CE) loss does not guarantee robust boundary for neural networks. CE loss sharpens the neural network at the decision boundary to achieve a lower loss, rather than pushing the boundary to a more robust position. A robust boundary should be kept in the middle of samples from different classes, thus maximizing the margins from the boundary to the samples. We think this is due to the fact that CE loss has no stationary point. In this paper, we propose a family of new losses, called stationary point (SP) loss, which has at least one stationary point on the correct classification side. We proved that robust boundary can be guaranteed by SP loss without losing much accuracy. With SP loss, larger perturbations are required to generate adversarial examples. We demonstrate that robustness is improved under a variety of adversarial attacks by applying SP loss. Moreover, robust boundary learned by SP loss also performs well on imbalanced datasets. \footnote{This work has been submitted to IEEE for possible publication. Copyright may be transferred without notice, after which this version may no longer be accessible.}
\end{abstract}


\IEEEpeerreviewmaketitle

\section{Introduction}

\IEEEPARstart{A}{rtificial}
intelligence based on deep neural network have achieved great success in speech recognition\cite{Hinton2012}, image classification\cite{Krizhevsky2012}, etc.
As neural networks are applying to various fields with security requirements, robustness has become an important attribute. However, \cite{Szegedy2013} and \cite{Kurakin2018a}, suggested that deep neural networks are vulnerable to adversarial examples. Adversarial examples are images that added subtle changes artificially on which the neural network makes mistakes but the changes can hardly be detected by human eyes.
 Specifically, adversarial examples are obtained within the neighborhood of samples in the training set, generally $L_{\infty}$ norm neighborhood. Since adversarial examples lead to misclassification, it is clearly that the neighborhood of training samples crosses the classification boundary. That is, the neural network is vulnerable to adversarial examples because its decision boundary is too closed to training samples, resulting in low robustness. For simplicity, we call the decision boundary which is too closed to training samples \emph{near-boundary}. In opposite to the \emph{near-boundary}, models with robust boundary maintains the prediction for an open set in the neighborhood of each training sample.

Categorical Cross-entropy (CE) loss is commonly used in image classification tasks, while generalized CE losses are also used in noisy labels, sound event classifications, and econometric models\cite{Zhang2018}, \cite{Deng2021}, \cite{Heckelei2008}, \cite{Kazemdehdashti2018}, \cite{Kurian2021}, etc. However, both \cite{Jacobsen2018} and \cite{Nar2019} draw the same conclusion that CE loss leads excessive invariance to predict features, i.e., classifiers make decisions relying on only a few highly predictive features. Specifically, CE loss will not continue to optimize the boundary once samples have been classified correctly based on a few features. We further empirically show that CE loss will only sharpen the inference function to achieve a lower loss rather than moving the decision boundary to a 'more robust' location once samples are classified correctly.

 Given the excessive invariance property, we consider \emph{near-boundary} is caused by CE loss. Specifically, we believe that CE loss dose not optimize to robust boundary is because it has no stationary point, which leads to no minimum of loss and thus the model keeps to increasing trainable weights to achieve lower loss. Thus, we propose a family of new losses with multiple stationary points to overcome the shortcoming.

The major contributions of this paper are as follow:
\begin{itemize}
  \item We identify that CE loss and focal loss sharp the confidence area and increase the weight of the fully connected layer of neural networks to reach a lower loss value rather than move the decision boundary to a more robust position.
  \item We propose a family of new losses, called stationary point (SP) loss, which has at least one stationary point on the correct classification side and we provide experimental evidence that SP loss learn a more robust boundary.
  \item We theoretically show that if a neural network model is trained by minimizing CE loss via gradient-based method, the model does not always convergence to robust boundary, but SP loss will.
  \item Several experiments of different neural network architectures on  multiple datasets with popular white box attack methods has been performed. The results show that SP loss can significantly enhance the network robustness without losing precision. Meanwhile, SP focal loss also works well on imbalanced datasets.
\end{itemize}

\section{RELATED WORK}

\begin{figure*}[!t]
\centering
\includegraphics[width=6.5in, height=4.3in]{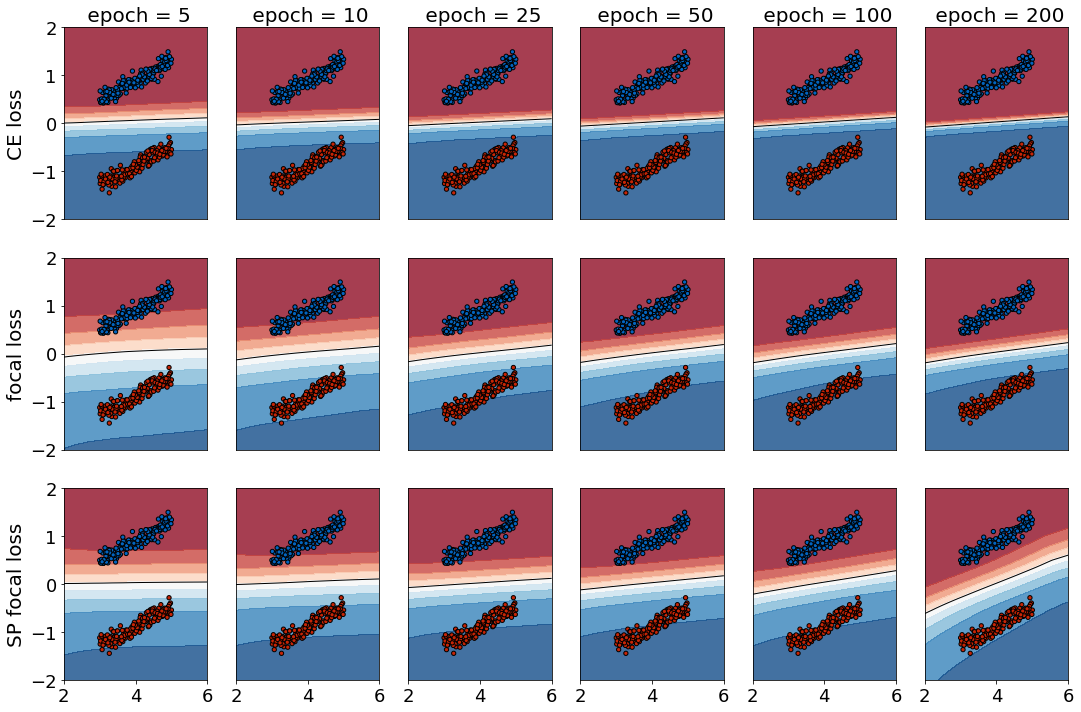}
\caption{Variation in decision boundaries and prediction confidences for CE loss, focal loss, and SP focal loss with increasing training epochs. The black solid line is the decision boundary. The background color is the prediction confidence.}
\label{fig:ce_confidence}
\end{figure*}

 Adversarial attacks have become one of the most important test of neural network robustness, notably the white box attacks the most compelling.
White box attacks require full access of the model, including the structure and parameters of each layer.
Advanced white box attack methods are Fast Gradient Sign Method (FGSM) \cite{Goodfellow2014}, Projected Gradient Descent (PGD)\cite{Madry2017}, the Carlini and Wagner (C\&W)\cite{Carlini2017}, etc.
\cite{Schmidt2018} argue that commonly used datasets can provide good accuracy, but not enough to provide a good robustness. Considering the difficulty of training robust classifiers in practice, \cite{Krizhevsky2009} hypothesized that, the difficulty may be the lack of training samples.

 In order to improve the robustness of neural networks, researchers have proposed many defensive methods. One of the most popular methods is the adversarial training. Adversarial training \cite{Madry2017}, \cite{Zhang2019}, \cite{Goodfellow2014}, \cite{Tramer2017} improves robustness in a data-augmented manner that generates additional training samples from adversarial methods. However, data generation based adversarial attacks is very expensive. Training with augmented data also degrades neural network performance compared to training with original dataset. There are many other defense methods, such as attention-based method \cite{Taghanaki2019}, \cite{Zoran2020}, \cite{Vaishnavi2020} and regularization methods \cite{Sokolic2017}, \cite{Cisse2017}, \cite{Kurakin2018}, \cite{Pang2019}. The defense methods make changes to the neural network, but do not guarantee generality. Meanwhile, with extra defensive components introduced, the training cost will increase as well.

A more effective and general methods is to design a new loss function to improve the robustness. Several kinds of loss or regularization based robust training method have been put forward. By applying potential characteristics regularization, convolutional neural networks (CNNs) are encouraged to study features between class separability and compactness within the class\cite{Pang2019a}, \cite{Mustafa2019}. Pang et al. \cite{Pang2019a} proposed Max-Mahalanobis center (MMC) loss, which studies identifiable features and looking for high density feature area. They first calculate the Max-Mahalanobis center for each class\cite{Pang2018}, then use the center loss to encourage features clustered around the center. However, the preset center $u_y^*$s for MMC are untrainable to avoid degradation. The choice of $u_y^*$s are crucial, which determines how features are clustered, but there are no guidelines for a good choice. Amid et al.\cite{Amid2019} proposed bi-tempered logistic loss by replacing the logarithm and the exponential functions of Softmax CE loss with their corresponding "tempered" version. Bi-tempered logistic loss improved the robustness to noisy data not to adversarial robustness.

\section{STATIONARY-POINT LOSS}
\subsection{A TOY EXAMPLE OF NEAR BOUNDARY}

We demonstrate \emph{near-boundary} of CE loss and even focal loss on a toy dataset. We created a two-class dataset (samples in blue and red) distributed as line segments. The samples are linearly separable with a vertical margin equals to 1.8. We trained a two-layer fully-connected neural network to classify the samples. As shown in the first row of Fig. \ref{fig:ce_confidence}, if CE loss is applied to the neural network, its decision boundary can separate, but, the decision boundary does not maximize margins between the boundary and samples on each side. Moreover, the region of low prediction confidence becomes narrower during training. It is because the CE loss keeps pushing the prediction probabilities of the sample towards 1 to reduce the loss. From another perspective of the view, the region of low prediction confidence means that the inference function is very sharp at the decision boundary. Thus, the prediction confidence no longer presents the true confidence of the model. The toy experiment demonstrates a behavior of the CE loss, i.e., when all samples are classified correctly, CE loss does not optimize the boundary to a robust position, but sharpen it to reduce the loss. It suggests that other forces are needed to push the CE loss to continue to optimize the boundary.

 Compared with CE loss, focal loss\cite{Lin2017} is designed for foreground and background imbalance during training process, namely the positive and negative samples imbalance. It reduces the penalty once the sample is correctly classified. As shown in the second row in Fig. \ref{fig:ce_confidence}, since correctly classified samples has less penalty, the region of low prediction confidence is much larger than the neural network trained under CE loss. However, the decision boundary is still a \emph{near-boundary} and the prediction confidence region is narrower as training epochs increasing.

 A robust decision boundary should maximize margins from the boundary to samples from both classes. In the third row of Fig. \ref{fig:ce_confidence}, with the same initialization, the decision boundary trained under our proposed SP loss is parallel to the line segments which are how training samples distributed.

\subsection{CE LOSS DOES NOT GUARANTEE TO CONVERGENCE TO A ROBUST BOUNDARY}
In the toy example, the CE loss dose not converge the model to the robust boundary. We analyzed this phenomenon from an optimization point of view. We found that \emph{near-boundary} can have lower loss than the robust boundary, so that model does not necessarily converge to the robust boundary.

\textbf{Notation}: Consider a classification problem. let $(X, y)\in\mathcal{D}$ be the training set, $X=\{\bm{x}_1, \bm{x}_2, ..., \bm{x}_N\}\in \mathbb{R}^{N \times d}$ is training sample, $N$ is the size of training set $\mathcal{D}$, $y\in \{1, 2, ..., S\}$ are corresponding labels. Let $f(\bm{x;\theta})$ represent a neural network model with $\bm{\theta}$ being the parameters. Denote $\hat{y}_k (k = 1,... , S)$ the output of the neural network before \emph{Softmax}, $\bm \xi$ is the predict of the network after \emph{Softmax}. Given a sample $x$, the output of the network after \emph{Softmax} is
\begin{align}
\xi_k = p(\bm \xi=k|x) = \frac{e^{\hat{y}_k}}{\sum_{k=1}^{S}e^{\hat{y}_k}},\quad k=1,...,S
\end{align}

\textbf{Theorem 1}  \textit{(CE loss does not necessarily converge to any particular boundary) For a neural network $f(\bm{x;\theta})$ trained with categorical CE loss $L_{cls}({\bm{\theta}})$ classifying a K-classed dataset $\mathcal{D}$. Let categorical CE loss as follows, }
\begin{align}
L_{cls}({\bm{\theta}}) = -\sum_{s=1}^{K}y_slog(\xi_s).
\end{align}
\textit{Using gradient-based method, we obtain the network $f(\bm{x;\theta})$. If $f(\bm{x;\theta})$ perfectly classified $\mathcal{D}$, i.e.}
\begin{align}
f(\bm{x}_i;\bm{\theta})_{y_i}>f(\bm{x}_i;\bm{\theta})_{k,k \neq y_i}, \forall (\bm{x}_i, y_i)\in \mathcal{D}
\end{align}
\textit{where $k \in \mathbb{N}, k\le K$, there is always another set of weights $\bm \theta^{'}$, such that $f(\bm{x;\theta ^{'}})$ classifies $\mathcal{D}$ perfectly as well, but with less categorical CE loss value, i.e.  $L_{cls}({\bm{\theta}})>L_{cls}({\bm{\theta} ^{'}})$.}

\begin{proof}
 Since $f(\bm{x;\theta})$ perfectly classified $\mathcal{D}$, there is no sample from $\mathcal{D}$ lying on the classification boundary. Consider the classification boundary for the class $k$,
\begin{align}
\partial S_k=\{\bm{x}:f(\bm{x;\theta})_k=f(\bm{x;\theta})_{k^{'}},k\neq k^{'}\}.
\end{align}
There exists a small enough offset $\delta_k \bm e$ such that $\{\bm{x}+\delta_k \bm{e}:f(\bm{x;\theta})_k=f(\bm{x; \theta})_{k^{'}}, k\neq k^{'}\}$ perfectly classifies the class $k$ as well, where $\delta_k>0$ and $\bm e$ is an arbitrary unit vector. Taking $\delta=min_k\delta_k$, we have
\begin{align}
\partial S_k^{'}=\{\bm{x}+\delta \bm{e}:f(\bm{x;\theta})_k=f(\bm{x;\theta})_{k^{'}},k\neq k^{'}\}
\end{align}
perfectly classified $\mathcal{D}$. $\partial S_k^{'}$ can also be written as
\begin{align}
\partial S_k^{'}=\{\bm{x}:f(\bm{x}-\delta \bm{e};\bm{\theta})_k=f(\bm{x}-\delta \bm{e};\bm{\theta})_{k^{'}},k\neq k^{'}\}.
\end{align}
Since $\bm e$ is an arbitrary unit vector, there always exists an $\hat{\bm{e}}$ and a set of weights $\bm{\theta^{'}}$, such that $f(\bm{x}-\delta \hat{\bm{e}};\bm{\theta})=f(\bm{x;\theta^{'}})$ by adjusting the weights in the first layer of the neural network. Hence, there exists $f(\bm{x; \theta ^{'}})$ perfectly classified $\mathcal{D}$. Because of the perfectly classification, the infimum of $L_{cls}({\bm{\theta^{'}}})$ is zero. So we can always enlarge all the weights in last layer of neural network by multiplying a constant to reach a lower loss without changing the decision boundaries. Thus, there is always another set of weights with less loss perfectly classified $\mathcal{D}$ as well.
\end{proof}

In Theorem 1, we demonstrate that  CE loss can convergence to zero at multiple decision boundaries. It implies that even a \emph{near boundary} can lead to the infimum loss. Thus, the robust boundary cannot be guaranteed by CE loss.

\subsection{STATIONARY POINT LOSS}
\begin{figure*}[!t]
\centering
\includegraphics[height=1.8in]{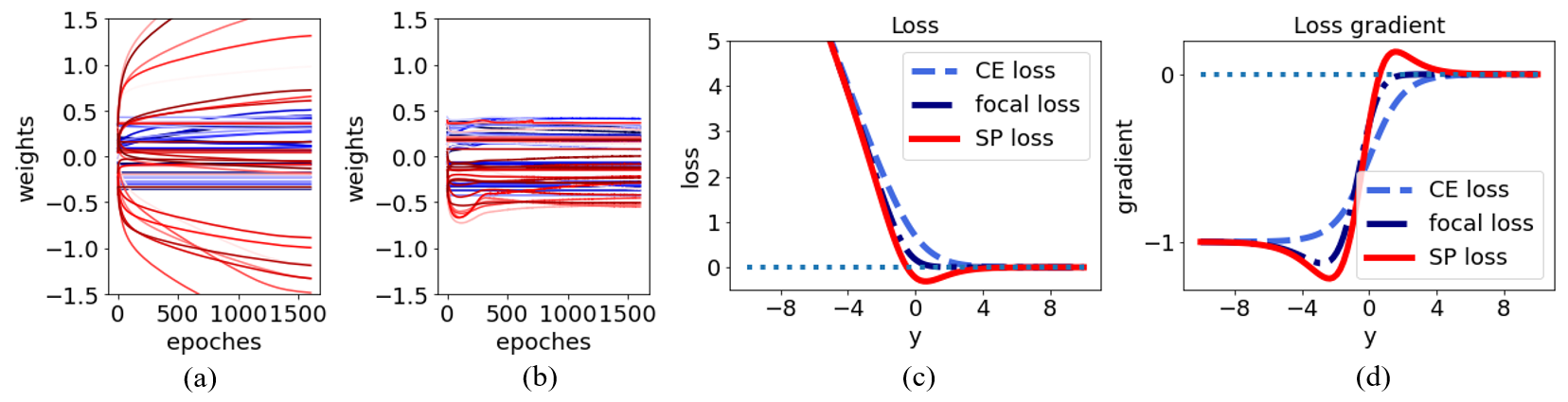}
\caption{ (a) Trend of the last layer weights during training under CE loss; (b) trend of the last layer weights during training under SP loss; (c) loss function of CE, focal, and SP loss; (d) loss gradient of CE, focal, and SP loss.}
\label{fig:new_loss}
\end{figure*}

From Theorem 1, the decision boundary obtained by CE loss is not guarantee to be robust. As the neural networks are vulnerable to adversarial attacks, decision boundaries learned from CE loss are \emph{near-boundaries}. On the other hand, to reach a lower loss, the neural network can increase the weights in the last layer, which will lead the region of low prediction confidence going narrower during training.

In Fig. $\ref{fig:new_loss}$ (a), we plotted weights in the last layer of the model trained under CE loss in the toy example. It shows that some if not all of the weights are diverged while training. The CE loss constantly sharpen the inference function at the decision boundary to achieve a lower loss and higher confidence scores. From athe differential equation point of view, we think the reason is that the CE loss has no stationary point. In other words, neural network parameters trained with CE loss is not converging.

 In Fig. $\ref{fig:new_loss}$ (c) and (d), we plotted the loss and loss gradient with respect to the neural network output before \emph{Softmax}. CE loss and focal loss both have positive losses which converge to zero as the neural network output tends to positive infinity. The only difference is that focal loss has almost zero loss value when the classification is correct, but CE loss still maintains a higher loss even the classification is correct. From their gradients, both CE loss and focal loss has negative gradient no matter the classification is correct or not. As there will be always negative gradients, neural network parameters may not converge.

To this end, we propose a new loss framework, namely Stationary Point (SP) loss, which has at least one stationary point by introducing a regularization term. The framework is as follows,
\begin{align}
SP(\bm \theta) : \mathcal{L} = L_{cls}(\bm \theta) + \eta \mathcal{R}(\bm \theta, \bm \xi)
\end{align}
where $L_{cls}(\bm \theta)$ is objective loss, $\mathcal{R}(\bm \theta, \bm \xi)$ is regularization term, which guarantees $SP(\bm \theta)$ has at least one stationary points at the correct classification, $\eta$ is a positive regularity coefficient. As the solid red line shown in Fig. $\ref{fig:new_loss}$ (c) and (d), SP loss has a minimum at the positive side of the neural network output which has zero gradient to ensure the local convergence of the weights. The Fig. $\ref{fig:new_loss}$ (b) shows the last layer weights of the neural network in the toy example trained under SP loss. Most of the weights converged comparing to the weights trained under CE loss.

It should be noted that for loss which has negative gradient about its input, we can always find a regularization term, which makes ${\rm d} SP(\bm \theta)/{\rm d}v|_{v=v_0}=0$. Specifically, let $f(\bm {x;\theta})$ be a neural network. The output of network $f(\bm {x;\theta})$ was firstly normalized in \emph{Softmax}, then it was fed into objective function, i.e. the loss, to optimize parameters. $L_{cls}$ represents the objective function, where \emph{Softmax(}$v$) is the input of $L_{cls}$. Let $L_{cls}$ be generalized CE loss with $L_{cls} \geq 0$. There always exists a suitable regularization term and coefficient $\mathcal{R}(v,\bm \theta), \eta >0$.

\subsubsection{A SPECIAL CASE: GRADIENT STARVATION REGULARIZATION}
Gradient starvation is a proclivity in neural networks discovered in \cite{Pezeshki2021}. It describes the phenomena that neural networks tend to be trained with only a few weights functioning. The other weights are "starved" with tiny backpropagation gradient during the gradient-based training. Pezeshki et al. proposed a regularization to ease the gradient starvation. We found that, gradient starvation is closely related to the \emph{near-boundary}. As shown in Fig. $\ref{fig:new_loss}$ (a), a few weights have dominant values comparing to other weights. Gradient starvation loss is a special case of SP loss. The gradient starvation loss is defined as
\begin{align}
\mathcal{L}(\bm \theta) = \bm{1} \cdot log[1 + exp(-\bm{Y\xi})] + \frac{\eta}{2} ||\bm \xi||^2
\end{align}
where $\bm{1}\cdot log[1 + exp(-\bm{Y\xi})]$ is the binary CE loss. If we rewrite gradient starvation loss in categorical CE loss, it becomes
\begin{align}\label{eq_cesp}
    L = -log(\bm \xi) + \eta (||\bm \xi|| ^2 + (||1-\bm \xi||)^2)
\end{align}
which is a stationary point loss when the regularization coefficient $\eta > 0.5$.

We denote the loss in \eqref{eq_cesp} with regularization coefficient $\eta > 0.5$ as SP CE loss. In a most simple scenario, we can show that SP CE loss converges to the most robust boundary, i.e., the midpoint of sample features.

\subsubsection{SP FOCAL LOSS}
we consider SP focal loss as
\begin{align}\label{FL}
FL(\bm \theta) :\mathcal{L}= -\alpha(1-\bm \xi)^ \gamma log(\bm \xi) + \eta (||\bm \xi|| ^2)
\end{align}
where $\bm \xi$ is the neural network output after \emph{Softmax}, $\eta$ is stationary-point coefficient.
The gradient of focal loss about neural network output before \emph{Softmax} is
\begin{align}
FL^{'}(\hat{y}_k):\mathcal{L}^{'} = \alpha (1-\bm \xi)^ \gamma (\gamma \bm \xi log(\bm \xi)-1+\bm \xi) + 2\eta \xi ^2(1- \xi)
\end{align}

We set $\alpha =0.25, \gamma = 2$ following the setting in \cite{Lin2017}. It can be easily shown that SP focal loss has one stationary point. Both the SP losses used in Fig. \ref{fig:ce_confidence} and Fig. \ref{fig:new_loss} are focal loss. Fig. \ref{fig:ce_confidence} shows that SP loss can reach a more robust boundary comparing with CE loss. Fig. \ref{fig:new_loss} shows that SP focal loss has better weight convergence.

\textbf{Lemma 1}  \textit{For binary classification neural network $f(\bm {x;\theta})$ trained with stationary-point loss $\mathcal{L}$. If there is only one sample of each class in the dataset, the classification boundary passes through the midpoint of the two sample features.}
\begin{proof}

Since $f(\bm {x;\theta})$ is trained with SP loss, there exists $z^{*}=[z_0^{*}, -z_0^{*}]^{\top}$ such that
 \begin{align}
 \notag
\frac{\partial \mathcal{L}}{\partial z}|_{z=z^{*}} = 0.
\end{align}
where $f(\bm {x;\theta})=\emph{Softmax}(z)$ and $z = [z_0, z_1]^{\top}$. Denoting the last fully-connected layer as $z_0=w_0 x_f + b_0$ and $z_1 = w_1 x_f + b_1$, we have
 \begin{align}
 \notag
w_0 x_f + b_0 &= z_0^{*} \\
\notag
w_1 x_f + b_1 &= -z_0^{*}
\end{align}
to minimize $\mathcal{L}$, where $x_f$ is the feature input of the fully-connected layer. Assuming the samples are $(x^{(0)}, 0)$ and $(x^{(1)}, 1)$, $\mathcal{L}$ will be minimized when
 \begin{align}
 \notag
w_0 x_f^{(0)} + b_0 &= z_0^{*} \\
\notag
w_1 x_f^{(0)} + b_1 &= -z_0^{*} \\
\notag
w_0 x_f^{(1)} + b_0 &= -z_0^{*} \\
\notag
w_1 x_f^{(1)} + b_1 &= z_0^{*}.
\end{align}

Then, we have
\begin{align}
\notag
w_0 = - w_1 &= \frac{2z_0^{*}}{x_f^{(0)}-x_f^{(1)}} \\
\notag
w_0 \bar{x}_f + b_0 &= w_1 \bar{x}_f + b_1 = 0 \\
\notag
 \bar{x}_f &= (x_f^{(0)}+x_f^{(1)})/2.
\end{align}

Thus, classification boundary passes through the midpoint of the two samples.
\end{proof}

\textbf{Corollary 1}  \textit{(SP CE loss convergences to the robust boundary) The neural network $f(\bm {x; \theta})$ trained with SP CE loss classifying a 2-classed dataset with training set $X=\{-1,1\}$ and label $y=\{1, 2\}$ converges to the robust decision boundary, i.e. the midpoint of two sample features $x=0$.}

Proof of Corollary 1 can be found in appendix A.

\textbf{Corollary 2}  \textit{(SP focal loss convergences to the robust boundary) The neural network $f(\bm {x; \theta})$ trained with SP focal loss classifying a 2-classed dataset with training set $X=\{-1,1\}$ and label $y=\{1, 2\}$ converges to the robust decision boundary, i.e. $x=0$.}

The proof is similar to Corollary 1.

\textbf{Theorem 2}  \textit{For multi-classification neural network $f(\bm {x;\theta})$ trained with stationary-point loss $\mathcal{L}_{sp}$ classifying a K-classed dataset $\mathcal{D}$. If $\mathcal{L}_{sp}$ reaches its lower bound on $\mathcal{D}$, then the classification boundary passes through the midpoint of any two sample features.}
\begin{proof}
Denote the SP loss for sample $(x, y)$ as $l_{sp}(x, y)$. Since SP loss has a stationary point, there exists $z^{(y)}\in\mathcal{R}^K$ with
$$z_k^{(y)}=\left \{
\begin{aligned}
&z_0, & if \quad k=y \\
&-z_0,& otherwise
\end{aligned}
\right.
$$
such that
 \begin{align}
 \notag
\frac{\partial \mathcal{L}_{sp}(x,y)}{\partial z_k}|_{z=z^{y}} = 0.
\end{align}
where $z$ is the neural network output before \emph{Softmax}, i.e. $f(\bm{x;\theta})= Softmax(z)$. Denote the last fully-connected layer as $z_k=  w^{(k)}u + b^{(k)}$, where $u(x)$ is the feature input of the fully-connected layer of sample $x$. Then, we have
 \begin{align}
 \notag
z_y &= w^{(y)}u + b^{(y)} = z_0 \\
\notag
z_k &=  w^{(k)}u + b^{(k)} = -z_0, \quad \forall k\neq y
\end{align}
to minimize $l_{sp}(x,y)$. Assume we take one sample from each of two different classes and name their feature-label pairs as $(u^{(i)}, y^{(i)})$ and $(u^{(j)}, y^{(j)})$. Since $\mathcal{L}_{sp}$ reaches its lower bound, both $(u^{(i)}, y^{(i)})$ and $(u^{(j)}, y^{(j)})$ reaches the stationary point. Thus, we have
 \begin{align}
 \notag
z_{y^{(i)}} &= w^{(y^{(i)})}u^{(i)} + b^{(y^{(i)})} = z_0 \\
\notag
z_{y^{(j)}} &= w^{(y^{(j)})}u^{(j)} + b^{(y^{(j)})} = -z_0 \\
\notag
z_y{^{(i)}} &= w^{(y^{(i)})}u^{(i)} + b^{(y^{(i)})} = z_0 \\
\notag
z_{y^{(j)}} &= w^{(y^{(j)})}u^{(j)} + b^{(y^{(j)})} = z_0.
\end{align}

The above equalities yields
 \begin{align}
 \notag
w^{(y^{(i)})}(u^{(i)}+u^{(j)})/2 + b^{(y^{(i)})} &= 0 \\
\notag
w^{(y^{(j)})}(u^{(i)}+ u^{(j)})/2 + b^{(y^{(j)})} &= 0
\end{align}
and
 \begin{align}
 \notag
w^{(y^{(i)})}(u^{(i)} - u^{(j)})/2 = 2z_0 \\
\notag
w^{(y^{(j)})}(u^{(i)} - u^{(j)}) = -2z_0.
\end{align}
Thus, classification boundary passes through the midpoint of $(u^{(i)}, y^{(i)})$ and $(u^{(j)}, y^{(j)})$.
\end{proof}

\textbf{Lemma 1} and \textbf{Theorem 2} proved that if $z^{*}=[z_0^{*}, -z_0^{*}]^{\top}$ is the stationary point of SP loss, then classification neural network passes through the midpoint of sample features. The following theorem shows that $z^{*}$ is one of the stationary points of SP loss.

\textbf{Theorem 3}  \textit{$z^{*}=[z_0^{*}, -z_0^{*}]^{\top}$ is one of the stationary points of SP loss.}
\begin{proof}
Consider binary classification, SP CE loss as follows,
 \begin{align}
\mathcal{L} = -log(\xi_k) + \eta ||\xi_k|| ^2, k=1, 2.
\end{align}
Then we take the derivative of $z_1, z_2$
\begin{align}
\frac{{\rm d}\xi_1}{{\rm d}z_1} = \xi_11-\xi_1)
\end{align}
\begin{align}
\frac{{\rm d}\xi_k}{{\rm d}z_2} = \xi_1(\xi_1-1).
\end{align}
Therefore,
\begin{align}
\notag
&\frac{{\rm d}\mathcal{L}}{{\rm d}z_1} = (\xi_1-1)(1-2\eta \xi_1^2) = 0 \\
\notag
&\xi_1 = \frac{1}{\sqrt{2\eta}}
\end{align}
${\rm d}\mathcal{L}/{\rm d}z_2$ is similar. Then we know that
\begin{align}\label{z2,-z1}
\notag
\xi_1 = \frac{e^{z_1}}{e^{z_1}+e^{z_2}}&=\frac{1}{1+e^{z_2-z_1}}=\frac{1}{\sqrt{2\eta}}\\
e^{z_2-z_1} &= \sqrt{2\eta} -1
\end{align}
$z_1=-z_2$ is one of solutions of \eqref{z2,-z1}.
\end{proof}

In above, we theoretically proved the effectiveness of SP loss for binary and multi-classification problems. In the following, we will conduct several experiments in classic networks and datasets to show the effectiveness of SP loss.

\section{EXPERIMENT}
In this section, we demonstrate several attractive instances of applying SP loss. We first validate SP loss on a linear inseparable toy dataset, which presents superior performance of SP loss visually. Next, we visualize the loss landscape of ResNet-18 to show that SP loss leads a more robust network. We also perform series experiments on MNIST, Fashion-MNIST\cite{Xiao2017}, CIFAR-10, CINIC-10\cite{Darlow2018}, Clothing1M\cite{Xiao2015} and VOC2007\cite{Everingham} datasets under white box attacks to validate the robustness improvement. Finally, an experiment on unbalanced dataset shows that SP loss compensentes for the shortcoming of imbalanced datasets to a certain extent.

\subsection{PERFORMANCE ON LINEAR INSEPARABLE DATASET}
\begin{figure}[!t]
\centering
\includegraphics[width=3.2in]{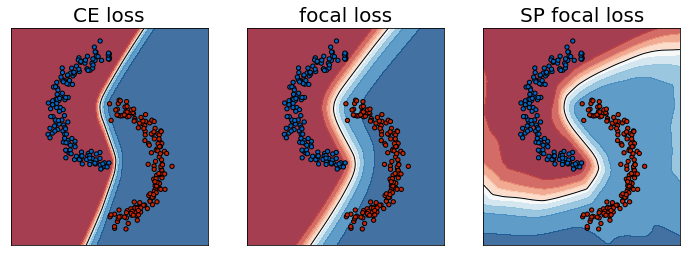}
\caption{Performance of models learned via CE loss, focal loss and SP focal loss, respectively, on a linear inseparable classification task.}
\label{fig:linear unseparable}
\end{figure}
In Fig. \ref{fig:ce_confidence}, we demonstrated that SP loss leads to a more robust decision boundary on a linear separable dataset. On linear inseparable dataset, models trained under SP loss are also more robust. We trained a two-layer full-connected neural network to classify a two-moons dataset with horizontal $margin=-0.01$. CE loss, focal loss and SP focal loss ($\eta=0.03$) are tested and compared. The color in background indicates corresponding confidence score. The brighter the area, the lower the confidence score. As Fig. $\ref{fig:linear unseparable}$ shows, the decision boundary learned by SP loss is closer to the middle position (robust boundary) than the decision boundaries learned by CE loss and focal loss. In addition, the decision boundary under SP loss follows the trend of the dataset while the other two don't.

 With robust boundary, it needs larger perturbation radius to generate adversarial examples. Intuitively, SP loss is more robust against adversarial attacks. We conducted several adversarial attack experiments in the following.

\subsection{ROBUSTNESS AGAINST ATTACKS}
\subsubsection{Loss landscape visualization}

\begin{figure}[!t]
\centering
\includegraphics[width=3.0in]{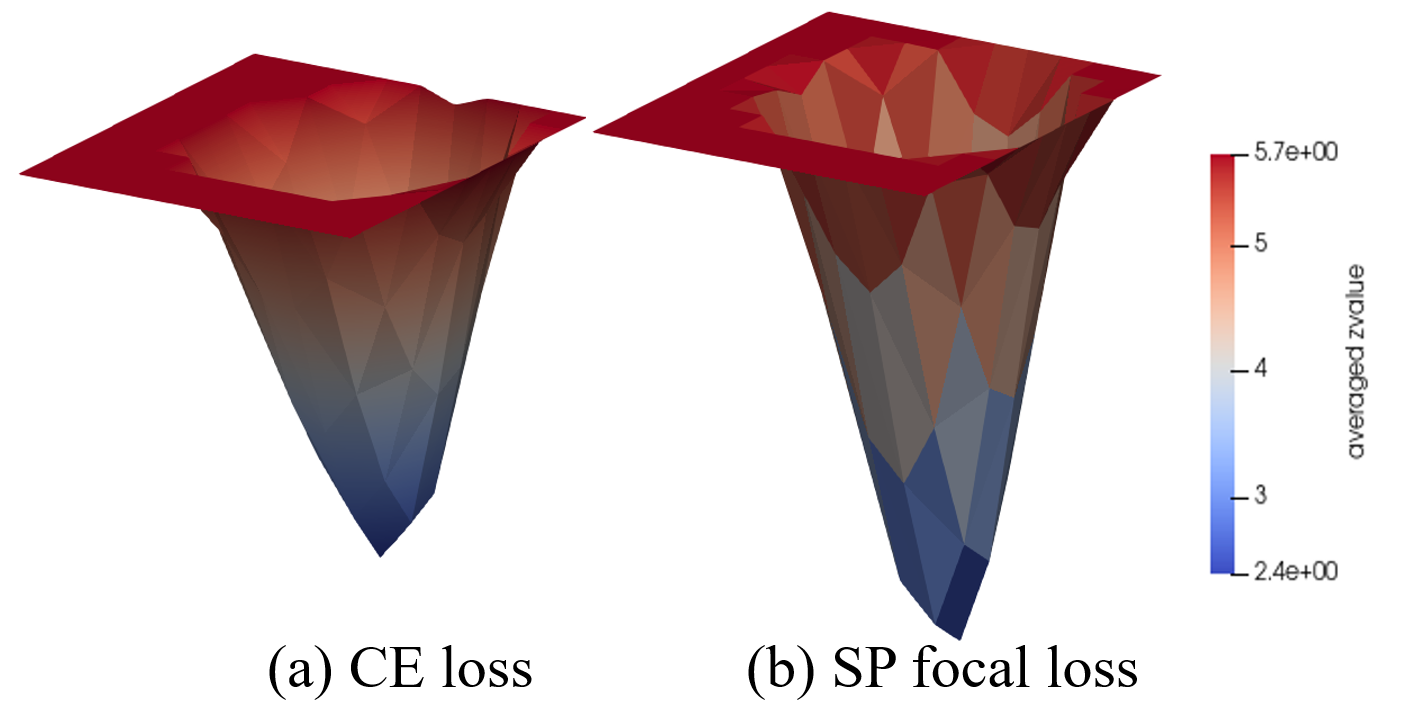}
\caption{Visualizing loss landscape.}
\label{fig:Landscape}
\end{figure}

In recent years, there have been many works investigating the relationship between loss landscape and robustness\cite{Keskar2016}. Prabhu etc.\cite{Prabhu2019} indicates that robust neural network has a sharper (or deeper) convex geometry.

Theorem 1 shows that CE loss has no minimum. Therefore, neural network trained with CE loss merely convergences to an approximate of the optimal in practice, i.e., there exists a small positive number $\alpha$ such that every $\bm \theta$ in weights set $S(\bm \theta)=\{\bm \theta | L_{\bm \theta}\le \alpha\}$ is considered as the critical point. Truncated by $\alpha$, the landscape of CE loss is a basin. In Fig. \ref{fig:Landscape}, we visualized the loss landscape by the method proposed by Li etc. \cite{Li2018}. The loss landscape of ResNet-18 with CE loss and SP loss are visualized respectively. Compared with CE loss landscape, SP loss landscape is sharper and deeper convex, which indicates SP loss is more robustness by\cite{Keskar2016}.

\subsubsection{Adversarial attack}
\begin{table*}
\renewcommand\arraystretch{1.5}
\caption{Classification accuracy(\%) comparison for ResNet-18, ResNet-34, and GoogleNet with CE loss and SP loss under several adversarial attack methods on Fashion-MNIST dataset.}
\label{adv_robust_fashionmnist}
\begin{center}
\begin{tabular}{ccccccc}
\hline
\multirow{2}{*}{\bf model}  &\multicolumn{2}{c}{\bf ResNet-18}  &\multicolumn{2}{c}{\bf ResNet-34} &\multicolumn{2}{c}{\bf GoogleNet} \\

                            &CE loss &SP focal loss &CE loss &SP focal loss &CE loss &SP focal loss
\\ \hline
Train Accuracy &100 & 100 &100 &100  &100 &99.97 \\

Test Accuracy &92.72 & 93.05 &93.01 &93.13  &93.38 &78.72 \\

FGSM  &93.55 & \bf{96.97} &93.68 &\bf{96.74}  &90.07 &\bf{92.84} \\

BIM   &62.03 &\bf{87.72} &62.78 &\bf{87.37} &54.50 &\bf{79.06} \\

PGD   &57.81 &\bf{77.54} &57.95 &\bf{76.54}   &49.10 &\bf{56.0} \\

UPGD  &24.11 &\bf{63.59} &26.0 &\bf{56.45} &12.42  &\bf{24.29}  \\
\hline
\end{tabular}
\end{center}
\end{table*}

\begin{table*}
\renewcommand\arraystretch{1.5}
\caption{Classification accuracy(\%) comparison for ResNet-18, ResNet-34, GoogleNet and DenseNet-121 with CE loss and SP loss under several adversarial attack methods on CIFAR-10 dataset.}
\label{adv_robust}
\begin{center}
\begin{tabular}{ccccccccc}
\hline
\multirow{2}{*}{\bf model}  &\multicolumn{2}{c}{\bf ResNet-18}  &\multicolumn{2}{c}{\bf ResNet-34} &\multicolumn{2}{c}{\bf GoogleNet} &\multicolumn{2}{c}{\bf DenseNet-121}\\

                            &CE loss &SP focal loss &CE loss &SP focal loss &CE loss &SP focal loss &CE loss &SP focal loss
\\ \hline
Train Accuracy &100 & 100 &100 &100  &100 &99.97 &100 &100 \\

Test Accuracy &76.18 & 76.06 &76.13 &77.08  &80.01 &78.72 &77.24 &79.46 \\

FGSM  &70.52 & \bf{90.56} &68.23 &\bf{88.40}  &64.82 &\bf{74.27} &50.37 &\bf{82.22} \\

BIM   &12.12 &\bf{71.19} &11.47 &\bf{68.86} &10.83 &\bf{44.05} &4.06 &\bf{54.92} \\

PGD   &0.14 &\bf{50.96} &0.11 &\bf{47.14}   &2.22 &\bf{23.66}  &0.90  &\bf{33.76}\\

UPGD  &0.12 &\bf{43.28} &0.1 &\bf{41.82} &2.27  &\bf{17.45}  &0.93 &\bf{23.07}  \\
\hline
\end{tabular}
\end{center}
\end{table*}

We consider several classic neural network structures, including ResNet-18, ResNet-34, GoogleNet and DenseNet-121, which trained in pytorch on  MNIST, Fashion-MNIST\cite{Xiao2017}, CIFAR-10, CINIC-10\cite{Darlow2018}, Clothing1M\cite{Xiao2015} and VOC2007\cite{Everingham} datasets, being adversarial attacked under white-attacks FGSM, BIM, PGD and UPGD. We applied Adam optimizer with initial learning rate 0.001 and trained for 50 epochs on MNIST and Fashion-MNIST, 100 epochs on CINIC-10, and Clothing-1M, 200 epochs on CIFAR-10 and 500 epochs for VOC2007. The most accurate models were saved and used for adversarial attack.

\begin{table*}
\renewcommand\arraystretch{1.5}
\caption{Adversarial accuracy(\%) comparison with other losses on CIFAR-10 dataset.}
\label{adv_loss}
\begin{center}
\begin{tabular}{ccccccc}
\hline
\multicolumn{1}{c}{\bf Perturbation}  &\multicolumn{1}{c}{\bf Attack} &\multicolumn{1}{c}{\bf SP focal loss}  &\multicolumn{1}{c}{\bf MMC}  &\multicolumn{1}{c}{\bf SCE} &\multicolumn{1}{c}{\bf Center loss} &\multicolumn{1}{c}{\bf L-DM}
\\  \hline
\multirow{2}{*}{$\epsilon=8/255$}   &$\text{PGD}_{10}$   &\bf{47.8}    &36.0 &3.7 &4.4 &19.8 \\ 

    &$\text{PGD}_{50}$    &\bf{39.7}    &24.8 &3.6 &4.3 &4.9 \\

\multirow{2}{*}{$\epsilon=16/255$}
   &$\text{PGD}_{10}$   &\bf{46.7}    &25.2 &2.9 &3.1 &11.0 \\ 

    &$\text{PGD}_{50}$    &\bf{26.7}    &17.5 &2.6 &2.9 &2.8 \\

\hline
\end{tabular}
\end{center}
\end{table*}

\begin{figure}[!t]
\centering
\includegraphics[width=3.2in]{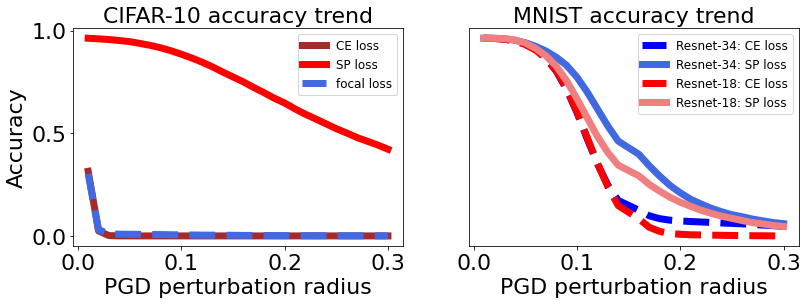}
\caption{Adversarial attack test in ResNet-18 and ResNet-34 on MNIST under PGD for $\epsilon=[0.01, 0.3]$.}
\label{fig:MNIST_pgd_radius_trend}
\end{figure}

For MNIST, Fashion-MNIST, CIFAR-10, Clothing-1M and VOC2007 datasets, adversarial settings are as follow. The max perturbation is $\epsilon=0.007$ for FGSM, $\epsilon=8/255$ for BIM. $\epsilon=8/255$ for PGD, and $\epsilon=4/255$ for UPGD. The number of iteration steps is set to 40 and the step size is set to $\alpha=1/255$. For CINIC-10, the max perturbation of PGD is $\epsilon=1/255$.

As shown in the second subplot of Fig. $\ref{fig:MNIST_pgd_radius_trend}$, CE loss can defense mild perturbation attacks on MNIST dataset. Both ResNet-18 and ResNet-34 have good accuracy when the PGD perturation radius is smaller than 0.1. As the perturbation radius increases, the classification accuracy gradually decreases. Models trained under SP loss remains higher accuracy than models trained under CE loss. SP loss based model is more robust on MNIST dataset.
SP loss has more precision improvement in Fashion-MNIST compared with MNIST. As Table $\ref{adv_robust_fashionmnist}$ shows, SP loss improved the accuracy by nearly 40\%.
 The results of adversarial attacks on CIFAR-10 are shown in Table $\ref{adv_robust}$.  Models trained under SP loss has comparable or even higher test accuracy comparing to models trained under CE loss. Note that adversarial training usually leads to accuracy decreasing but SP loss dosen't. SP loss improved the accurate of all tested attack methods for at least 20\%. For ResNet-18 and ResNet-34, SP loss improved the accuracy by around 50\% under BIM and PGD attacks.

  We future analyzed the accuracy of ResNet-18 on CIFAR-10 under PGD with perturbation radius from from 0.001 to 0.3.
As the first subplot of Fig. $\ref{fig:MNIST_pgd_radius_trend}$ shows, SP loss remarkably improved the accuracy under adversarial attack. The accuracy under adversarial attack. The accuracy under SP loss gradually decreased as the perturbation radius increase. However, the accuracy under CE loss and focal loss dramatically decreased to zero influenced by a tiny perturbation. Thus, on CIFAR-10, SP loss is more stable to the increasing perturbation radius.

\begin{table*}
\renewcommand\arraystretch{1.5}
\caption{Classification accuracy(\%) comparison for ResNet-18, ResNet-34, GoogleNet and DenseNet-121 with CE loss and SP loss under several adversarial attack methods on CINIC-10 dataset.}
\label{adv_robust_cinic10}
\begin{center}
\begin{tabular}{ccccccccc}
\hline
\multirow{2}{*}{\bf model}  &\multicolumn{2}{c}{\bf ResNet-18}  &\multicolumn{2}{c}{\bf ResNet-34} &\multicolumn{2}{c}{\bf GoogleNet} &\multicolumn{2}{c}{\bf DenseNet-121}\\

                            &CE loss &SP focal loss &CE loss &SP focal loss &CE loss &SP focal loss &CE loss &SP focal loss
\\ \hline
Train Accuracy &99.31 & 89.30 &99.08 &99.10  &98.95 &99.65 &99.08 &98.94 \\

Test Accuracy &15.80 & 14.09 &16.41 &16.47  &46.05 &51.91 &16.94 &17.63 \\

FGSM  &9.40 & \bf{25.70} &32.56 &\bf{39.50}  &9.27 &\bf{50.40} &21.00 &\bf{41.60} \\

BIM   &8.70 &\bf{59.0} &6.40&\bf{13.5} &0 &\bf{10.4} &1.58 &\bf{17.42} \\

PGD   &9.00 &\bf{16.10} &16.20 &\bf{19.5}   &0.10 &\bf{10.29}  &6.65 &\bf{17.70}\\

UPGD  &8.50 &\bf{10.02} &0.0 &\bf{2.89} &0  &\bf{0.035}  &0.78 &\bf{1.50}  \\
\hline
\end{tabular}
\end{center}
\end{table*}

\begin{table*}
\renewcommand\arraystretch{1.5}
\caption{Adversarial accuracy(\%) comparison with other losses, Clothing-1M dataset.}
\label{adv_robust_Clothing1M}
\begin{center}
\begin{tabular}{ccccccccc}
\hline
\multirow{2}{*}{\bf model}  &\multicolumn{2}{c}{\bf ResNet-18}  &\multicolumn{2}{c}{\bf ResNet-34} &\multicolumn{2}{c}{\bf GoogleNet} &\multicolumn{2}{c}{\bf DenseNet-121}\\

                            &CE loss &SP focal loss &CE loss &SP focal loss &CE loss &SP focal loss &CE loss &SP focal loss
\\ \hline
Train Accuracy &99.88 & 99.58 &99.86 &99.49  &98.40 &99.10 &98.98 &98.65 \\

Test Accuracy &65.23 & 66.79 &67.04 &67.00  &67.95 &64.27 &69.69 &69.84 \\

FGSM  &1.90 & \bf{11.55} &1.60 &\bf{8.70}  &1.30 &\bf{7.00} &1.34 &\bf{7.99} \\

BIM   &0.002 &\bf{7.00} &0.004 &\bf{4.66} &0.004 &\bf{1.00} &0 &\bf{3.00} \\

PGD   &0 &\bf{5.40} &0 &\bf{1.85}   &0.002 &\bf{0.48}  &0 &\bf{0.60}\\

UPGD  &0 &\bf{0.014} &0.002 &\bf{0.004} &0  &\bf{0.008}  &0 &\bf{0.006}  \\
\hline
\end{tabular}
\end{center}
\end{table*}

\begin{table*}
\renewcommand\arraystretch{1.5}
\caption{Classification accuracy(\%) comparison for ResNet-18, ResNet-34, GoogleNet and DenseNet-121 with CE loss and SP loss under several adversarial attack methods on VOC2007 dataset.}
\label{adv_robust_voc2007}
\begin{center}
\begin{tabular}{ccccccccc}
\hline
\multirow{2}{*}{\bf model}  &\multicolumn{2}{c}{\bf ResNet-18}  &\multicolumn{2}{c}{\bf ResNet-34} &\multicolumn{2}{c}{\bf GoogleNet} &\multicolumn{2}{c}{\bf DenseNet-121}\\

                            &CE loss &SP focal loss &CE loss &SP focal loss &CE loss &SP focal loss &CE loss &SP focal loss
\\ \hline
Train Accuracy &90.20 & 91.92 &88.63 &99.49  &74.45 &77.61 &90.32 &91.88 \\

Test Accuracy &63.58 & 64.46 &65.00 &67.0  &55.34 &56.87 &65.06 &66.53 \\

FGSM  &33.00 & \bf{51.00} &4.60 &\bf{6.00}  &9.10 &\bf{14.60} &6.00 &\bf{23.50} \\

BIM   &2.44 &\bf{7.00} &1.60&\bf{4.10} &0.40 &\bf{1.40} &0 &\bf{0.50} \\

PGD   &1.30 &\bf{3.00} &1.00 &\bf{4.00}   &0 &\bf{0.80}  &0.04 &\bf{0.30}\\

UPGD  &0.08 &\bf{0.16} &0 &\bf{1.00} &0  &0  &0 &\bf{0.12}  \\
\hline
\end{tabular}
\end{center}
\end{table*}

We compared SP loss with other losses which are proposed for robust models, including MMC\cite{Pang2019a}, SCE \cite{He2016a}, Center loss \cite{Wen2016a}, L-DM \cite{Wan2018} under PGD attack. The PGD attack with maximum perturbation $\epsilon=8/255, 16/255$ and steps=10, 50 are compared. Since the untrainable parameter $\mu_y^{*}$ is preset, and the code is unpublished, the reimplementation of other methods are following the original papers. As Table $\ref{adv_loss}$ shows, SP loss achieved at least 10\% highest adversarial accuracy than comparing methods.

At last, we conducted several experiments on large datasets, including CINIC-10\cite{Darlow2018}, Clothing-1M\cite{Xiao2015} and VOC2007\cite{Everingham}.
CINIC-10 has a total of 270000 images, 4.5 times larger than CIFAR-10. Apart from CIFAR-10 dataset, CINIC-10 also contains images from a selection of ImageNet. The image size of CINIC-10 is $32\times 32$ pixels. Table $\ref{adv_robust_cinic10}$ shows that SP loss improved the adversarial accuracy up to $41.13\%$ under FGSM, $50.3\%$ under BIM and $11.05\%$ under PGD. It shows that SP loss is equally effective for large datasets.
Clothing-1M contains 1M clothing images in 14 classes. It is a dataset with noisy labels. In this paper we only use the dataset with clean labels, which contains 50k images, including 14k images for training and validation, respectively. As Table $\ref{adv_robust_Clothing1M}$ shows, Clothing-1M is more vulnerable to adversarial attack. SP loss achieved higher adversarial accuracy than CE loss as well.
VOC2007 dataset has more complicated image background. The image size is various, generally for $500\times375$. Table $\ref{adv_robust_voc2007}$ shows that SP loss could adapt to various datasets and various images sizes.

\subsection{DATA ROBUSTNESS}
\begin{figure}[!t]
\centering
\includegraphics[width=3.2in]{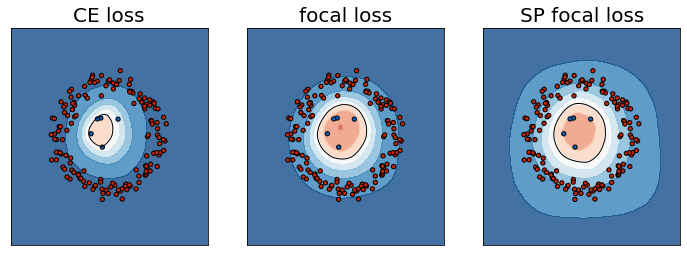}
\caption{Decision boundary learned by CE loss, focal loss and MS loss, respectively on imbalanced dataset which is lacking 97\% data points.}
\label{fig:twocircle1.5_0.97}
\end{figure}

Imbalanced data is a classic problem in classification tasks, which requires new understandings, principles, algorithms, and tools to transform vast amounts of raw data efficiently into information and knowledge representation \cite{He2009}. In short, imbalanced data problem is the number of samples under each category varies greatly in datasets. Imbalanced data usually results in high training accuracy but low testing accuracy of NNs model. In this case, it is obvious that the classifier is invalid. So it is very important to improve imbalanced data problem.
Due to epistemic uncertainty, we may never collect a complete datasets, and the work of collecting data could be costly. It would be a cost effective way to improve the robustness problem through the loss function.

Consider a binary classification task. Two classes are equal in number under the natural condition. For some reason, we get some imbalanced data of two classes. We conduct a experiment on a two layers full-connection neural network, to compare the boundary when the network trained on the datasets in lacking 10\%, 30\%, 50\% and 90\% respectively. The experiment results lacking 97\% are exhibited in Fig. $\ref{fig:twocircle1.5_0.97}$, other results can be found in appendix B.
From the Fig. $\ref{fig:twocircle1.5_0.97}$, decision boundary CE loss learned is far away from the larger volume data, which means the large classification probability of the outer data, and small classification probability of the inner data. Compared with CE loss, classification boundary learned by SP focal loss is closer to the medium location. This reduces the effect of category imbalance on the classification probability.
\subsection{THE DIFFERENCE BETWEEN SP LOSS AND L2 REGULARIZATION}
L2 regularization is proposed for over-fitting problem. Over-fitting is a shortcoming of machine learning models that has often been criticized. Over-fitting means that the algorithm performs well on the training set, but does not perform well on the test set, resulting in poor generalization. Intuitively, large numerical weight vectors are severely penalized by L2 regularization. The regularization encourage parameter to be smaller values. If the parameter $w$ is less than 1, then the $w^2$ will be smaller.
\begin{figure}[t]
\centering
\includegraphics[width=3.5in]{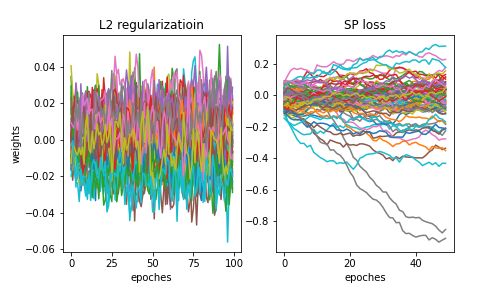}
\caption{Visualizing loss landscape}
\label{fig:weight_vis}
\end{figure}

To compare the difference between SP loss and L2 regularization, we classified the MNIST dataset on a two-layer convolutional neural network. The weights of the last layer of the two-layer convolutional neural network are visualized. As Fig. \ref{fig:weight_vis} shows, L2 regularization forced the weights of the CNN around zero. But SP focal loss driven the weights limit to less than 1 and not diverge.
\section{CONCLUSION}

In this paper, we proposed a family of new losses, called stationary point (SP) loss, which introduce additional stationary points to improve the model robustness. Firstly, we conducted a toy experiment to demonstrate that CE loss and focal loss stop optimizing the boundary to a more robust position once all samples are classified correctly, but sharpen the confidence area and increasing the weight of the last layer to reach a lower loss. Next, we showed that neural networks trained by CE loss cannot guarantee robust decision boundary theoretically. And we proved that robust boundary can be guaranteed by SP loss without losing much accuracy. Then, we validated the effectiveness of SP loss with an experiment of two-layers neural network on toy dataset. Finally, we conducted several experiments to show that SP losses improve the model robustness by visualizing loss landscape and adversarial attack under several different attack methods. SP focal loss also perform well on imbalanced data.

In our experiments, we found that SP focal loss performs better on those datasets and models which generalzed well. It has better performance on those datasets has less categories. In future work, it is necessary to design stationary point losses suitable for classification problems with large number of classes. A theoretical analysis for SP loss on general learning problem is also important.

\appendices
\section{}
\textbf{Theorem 1}  \textit{(CE loss does not necessarily converge to any particular boundary) For a neural network $f(\bm{x;\theta})$ trained with categorical CE loss $L_{cls}({\bm{\theta}})$ classifying a K-classed dataset $\mathcal{D}$. Let categorical CE loss as follows, }
\begin{align}
L_{cls}({\bm{\theta}}) = -\sum_{s=1}^{K}y_slog(\xi_s).
\end{align}
\textit{Using gradient-based method, we obtain the network $f(\bm{x;\theta})$. If $f(\bm{x;\theta})$ perfectly classified $\mathcal{D}$, i.e.}
\begin{align}
f(\bm{x}_i;\bm{\theta})_{y_i}>f(\bm{x}_i;\bm{\theta})_{k,k \neq y_i}, \forall (\bm{x}_i, y_i)\in \mathcal{D}
\end{align}
\textit{where $k \in \mathbb{N}, k\le K$, there is always another set of weights $\bm \theta^{'}$, such that $f(\bm{x;\theta ^{'}})$ classifies $\mathcal{D}$ perfectly as well, but with less categorical CE loss value, i.e.  $L_{cls}({\bm{\theta}})>L_{cls}({\bm{\theta} ^{'}})$.}

\begin{proof}

 We offer another proof of \textbf{Theorem 1}. First, we prove \textbf{theorem 1} in 1 dimension, it is same in multi dimension. Assuming we have two data points, -1, 1, then we know that the decision boundary is $x=0$. Now we assume network trained by CE loss got the decision boundary in $x=0$ in some weights $w_{\eta}$ in the network training process, in which CE loss value is $\eta$.
CE loss is
$$L(p)=-log(p)$$.
The derivation is
$$L^{'}(p)= -\frac{1}{p}$$
and
$$L^{'}(p)<0$$
is always true. We assert that there must be another decision boundary got by current state. Assume another decision boundary corresponding weights is $w_{\xi}$, the CE loss value is $\xi$. Then we have

\begin{align}
    w_{\xi} = w_{\eta} + \frac{\alpha}{p}
\end{align}
in which $\alpha$ is learning rate. Assume $x>0$ is positive sample, negative sample is also true.
\begin{align}
    \eta = -log \frac{1}{1+e^{1-2(w_{\eta} x + b)}}
\end{align}

\begin{align}
    \xi =  -log \frac{1}{1+e^{1-2(w_{\eta} x + b)-2\frac{\alpha x}{p}}}
\end{align}
Due to

\begin{align}
    exp^{1-2(w_{\eta} x + b)} > e^{1-2(w_{\eta} x + b)-2\frac{\alpha x}{p}}
\end{align}

therefore
\begin{align}
    -log \frac{1}{1+e^{1-2(w_{\eta} x + b)}} > -log \frac{1}{1+e^{1-2(w_{\eta} x + b)-2\frac{\alpha x}{p}}}.
\end{align}
So
\begin{align}
    \eta > \xi
\end{align}

\textbf{Proof--High dimension}: Considering a $n$ classes classification task. Assume input $x\in\mathcal{R}^d$, $f(x,\theta)$ is a L-layer full convolutional network, the output of the network is
\begin{align}
    y_K = \sum_{l=1}^{K-1}w_l^Ty_{l-1} + b_l
\end{align}
where $y_{l-1}$ is the output of $l-1$ layer, $y_0$ is the input x.
Assume that $f(x,\theta)$ represents a robust boundary with parameters $w_\xi$,
Then after $softmax$,
\begin{align}
    \xi _k = \frac{exp^{\sum_{l=1}^{L-1}w_l^Ty_{l-1} + b_l + w_{Lk} + b_{Lk}}}{\sum_{j=1}^{n}exp^{\sum_{l=1}^{L-1}w_l^Ty_{l-1} + b_l + w_{Lj} + b_{Lj}}}
\end{align}
Optimized by gradient-based method, there is another parameters $w_\eta$ satisfy
\begin{align}
w_\eta = w_\xi + \alpha \bigtriangledown_{ w_\xi} L_{ce}
\end{align}
where $\alpha$ is learning rate. Simply, note $w_\xi$ to $w$
The after $softmax$ output with parameters $w_\eta$ is
\begin{align}
    \eta _k &= \frac{e^{\sum_{l=1}^{L-1}w_{lk}^Ty_{l-1} + b_{lk} + \frac{\alpha y_{l-1}}{\bigtriangledown_{ w_l} L_{ce}} + w_{Lk} + b_{Lk} + \frac{\alpha y_{L-1}}{\bigtriangledown_{w_k} L_{ce}}}}{\sum_{j=1}^{n}e^{\sum_{l=1}^{L-1}w_{lj}^Ty_{l-1} + b_{lj} + \frac{\alpha y_{l-1}}{\bigtriangledown_{ w_l} L_{ce}} + w_{Lj} + b_{Lj}+ \frac{\alpha y_{L-1}}{\bigtriangledown_{w_k} L_{ce}}}}\\
    &= \frac{e^{\sum_{l=1}^{L-1}w_l^Ty_{l-1} + b_l + w_{Lk} + b_{Lk}}e^{Grad \_up}}{\sum_{j=1}^{n}e^{\sum_{l=1}^{L-1}w_l^Ty_{l-1} + b_l + w_{Lj} + b_{Lj}}e^{Grad \_down}}\\
    &=\frac{e^{\sum_{l=1}^{L-1}w_l^Ty_{l-1} + b_l + w_{Lk} + b_{Lk}}}{\sum_{j=1}^{n}e^{\sum_{l=1}^{L-1}w_l^Ty_{l-1} + b_l + w_{Lj} + b_{Lj}}e^{Grad \_down-Grad \_up}}
\end{align}
${Grad\_down-Grad\_up}$ denotes all the variation throughout the model.
Because
\begin{align}
 exp^{Grad_part_down-Grad_part_up} > 0
\end{align}
\begin{align}
\eta_k < \xi_k
\end{align}
\end{proof}

\textbf{Corollary 1}  \textit{(SP CE loss convergences to the robust boundary) The neural network $f(\bm {x; \theta})$ trained with SP CE loss classifying a 2-classed dataset with training set $X=\{-1,1\}$ and label $y=\{1, 2\}$ converges to the robust decision boundary, i.e. the midpoint of two sample features $x=0$.}
\begin{proof}
 For binary classification task, insert training examples $X$ into SP loss. Let

    \begin{align}
        \xi = \frac{1}{1+e^{1-2(w+b)}}
    \end{align}

\begin{align}
    \gamma = \frac{1}{1+e^{1-2(b-w)}}
\end{align}

For binary classification task, SP regularization term $||\bm \xi||^2=\xi ^2+(1-\xi)^2$, where $\bm \xi = (\xi, 1-\xi)$. Therefore, the SP loss is
\begin{align}
    - log \xi + \eta {\xi}^2+\eta (1-\xi )^2- log(1-\gamma )+\eta {\gamma}^2 + \eta (1-\gamma)^2 = L
\end{align}
$\eta$ is two-stationary coefficient.

As follows, we separate L to two parts,
\begin{align}
\begin{split}
   &L1 := -log \xi + \eta {\xi}^2+\eta (1-\xi )^2 \\
    &L2 := -log(1-\gamma )+\eta {\gamma}^2 + \eta (1-\gamma)*2
\end{split}
\end{align}

differentiate them separately,
  \begin{align}\label{I}
        L1_w ^{'} &= -\frac{1}{\xi} {\xi}_w ^{'}+2\eta \xi {\xi}_w ^{'} + 2\eta (1-\xi)(-1) {\xi}_w ^{'} \\
                    \notag
                  &= (-2)(1-\xi)[-1+2\eta {\xi}^2 - 2\eta \xi (1-\xi)]
    \end{align}

\begin{align}\label{II}
L2_w ^{'} &= -\frac{1}{1-\gamma}(-1){\gamma}_w^{'}+2\eta \gamma {\gamma}_w^{'}+2\eta(1-\gamma)*(-1){\gamma}_w^{'}\\
          \notag
          &= 2\gamma(1+2\eta \gamma(1-\gamma)-2\eta (1-\gamma)^2)
\end{align}

\begin{align}\label{III}
L1_b^{'} &= -\frac{1}{p} {\xi}_b ^{'}+2\eta \xi {\xi}_b ^{'} + 2\eta (1-\xi)(-1) {\xi}_b ^{'} \\
         \notag
         &= (-2)(1-\xi)[-1+2\eta{\xi}^2-2\eta \xi (1-\xi)]
\end{align}
\begin{align}\label{IV}
L2_b ^{'} &= -\frac{1}{1-\gamma}(-1){\gamma}_b^{'}+2\eta \gamma {\gamma}_b^{'}+2\eta(1-\gamma)*(-1){\gamma}_b^{'}\\
         \notag
         &= -2\gamma(1+2\eta \gamma(1-\gamma)-2\eta (1-\gamma)^2)
\end{align}

From formula $(\ref{I}), (\ref{II})$, differentiate $w$ we have
\begin{align}\label{der_w}
    (-2)(1-\xi)[-1+2\eta {\xi}^2 - 2\eta \xi (1-\xi)] \\
    \notag
    +2\gamma(1+2\eta \gamma(1-\gamma)-2\eta (1-\gamma)^2)=0
\end{align}
Similarly, differentiate $b$ we have
\begin{align}\label{der_b}
    (-2)(1-\xi)[-1+2\eta{\xi}^2-2\eta \xi (1-\xi)]\\
    \notag
    -2\gamma(1+2\eta \gamma(1-\gamma)-2\eta (1-\gamma)^2)=0
\end{align}

incorporate formulas $(\eqref{der_w}), (\eqref{der_b})$, we get
\begin{align}\label{s1}
    {\xi}_1=\frac{\sqrt{\eta (4+\eta)}+\eta}{4\eta}
\end{align}
\begin{align}\label{s2}
    {\xi}_2=\eta - \frac{\sqrt{\eta (4+\eta)}}{4\eta},(del)
\end{align}
\begin{align}\label{s3}
    {\gamma}_1=3 \eta - \frac{\sqrt{\eta (4+\eta)}}{4\eta}
\end{align}
\begin{align}\label{s4}
    {\gamma}_2=\frac{\sqrt{\eta (4+\eta)}+3 \eta}{4\eta}
\end{align}
then incorporate formula $\eqref{s1}, \eqref{s3}$we get $b=0.5$, so the decision boundary contains point $x=0$, so is the center of the dataset.
\end{proof}

\section{}

This section presents more experimental results from unbalanced datasets.
\begin{figure}[!t]
\centering
\includegraphics[width=3.2in]{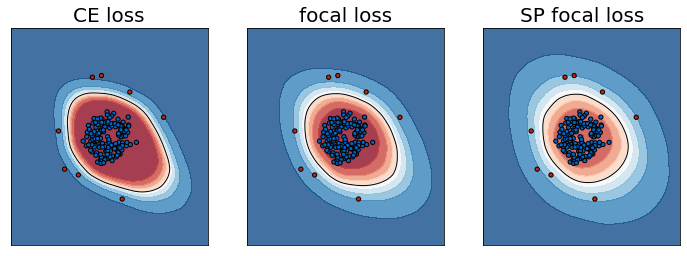}
\caption{Imbalanced dataset in lacking 95\% outside data points.}
\label{fig:im_01}
\end{figure}

\begin{figure}[!t]
\centering
\includegraphics[width=3.2in]{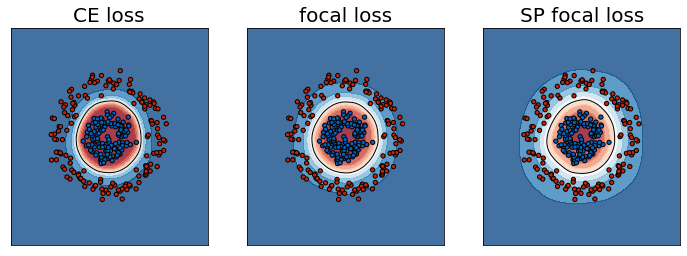}
\caption{Imbalanced dataset in lacking 10\% data points.}
\label{fig:im_01}
\end{figure}

\begin{figure}[!t]
\centering
\includegraphics[width=3.2in]{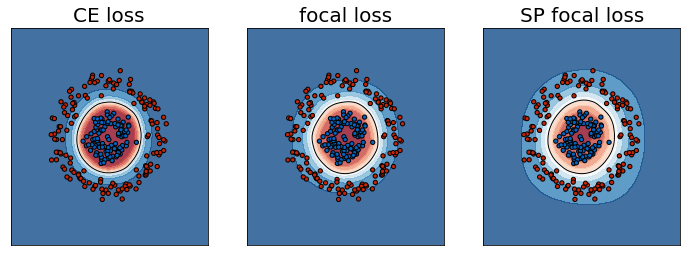}
\caption{Imbalanced dataset in lacking 30\% data points.}
\label{fig:im_03}
\end{figure}

\begin{figure}[!t]
\centering
\includegraphics[width=3.2in]{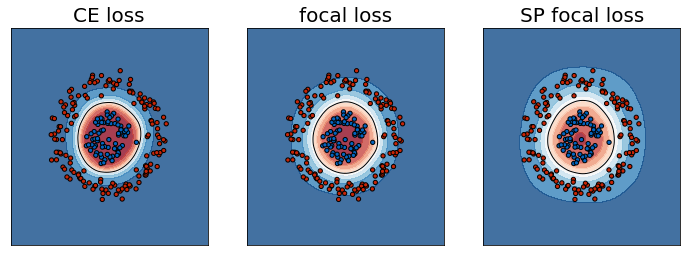}
\caption{Imbalanced dataset in lacking 50\% data points.}
\label{fig:im_05}
\end{figure}

\begin{figure}[!t]
\centering
\includegraphics[width=3.2in]{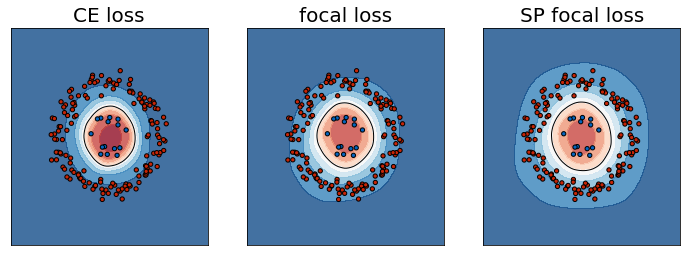}
\caption{Imbalanced dataset in lacking 90\% data points.}
\label{fig:im_09}
\end{figure}

\cite{*}
\bibliographystyle{IEEEtran}
\bibliography{SPabrv, SPexample}

\begin{thebibliography}{10}
\providecommand{\url}[1]{#1}
\csname url@samestyle\endcsname
\providecommand{\newblock}{\relax}
\providecommand{\bibinfo}[2]{#2}
\providecommand{\BIBentrySTDinterwordspacing}{\spaceskip=0pt\relax}
\providecommand{\BIBentryALTinterwordstretchfactor}{4}
\providecommand{\BIBentryALTinterwordspacing}{\spaceskip=\fontdimen2\font plus
\BIBentryALTinterwordstretchfactor\fontdimen3\font minus
  \fontdimen4\font\relax}
\providecommand{\BIBforeignlanguage}[2]{{%
\expandafter\ifx\csname l@#1\endcsname\relax
\typeout{** WARNING: IEEEtran.bst: No hyphenation pattern has been}%
\typeout{** loaded for the language `#1'. Using the pattern for}%
\typeout{** the default language instead.}%
\else
\language=\csname l@#1\endcsname
\fi
#2}}
\providecommand{\BIBdecl}{\relax}
\BIBdecl

\bibitem{Hinton2012}
G.~Hinton, L.~Deng, D.~Yu, G.~E. Dahl, A.-r. Mohamed, N.~Jaitly, A.~Senior,
  V.~Vanhoucke, P.~Nguyen, T.~N. Sainath \emph{et~al.}, ``Deep neural networks
  for acoustic modeling in speech recognition: The shared views of four
  research groups,'' \emph{IEEE Signal processing magazine}, vol.~29, no.~6,
  pp. 82--97, 2012.

\bibitem{Krizhevsky2012}
\BIBentryALTinterwordspacing
A.~Krizhevsky, I.~Sutskever, and G.~E. Hinton, ``Imagenet classification with
  deep convolutional neural networks,'' in \emph{Advances in Neural Information
  Processing Systems}, F.~Pereira, C.~Burges, L.~Bottou, and K.~Weinberger,
  Eds., vol.~25.\hskip 1em plus 0.5em minus 0.4em\relax Curran Associates,
  Inc., 2012. [Online]. Available:
  \url{https://proceedings.neurips.cc/paper/2012/file/c399862d3b9d6b76c8436e924a68c45b-Paper.pdf}
\BIBentrySTDinterwordspacing

\bibitem{Szegedy2013}
C.~Szegedy, W.~Zaremba, I.~Sutskever, J.~Bruna, D.~Erhan, I.~Goodfellow, and
  R.~Fergus, ``Intriguing properties of neural networks,'' 2013.

\bibitem{Kurakin2018a}
A.~Kurakin, I.~J. Goodfellow, and S.~Bengio, ``Adversarial examples in the
  physical world,'' in \emph{Artificial intelligence safety and
  security}.\hskip 1em plus 0.5em minus 0.4em\relax Chapman and Hall/CRC, 2018,
  pp. 99--112.

\bibitem{Zhang2018}
Z.~Zhang and M.~Sabuncu, ``Generalized cross entropy loss for training deep
  neural networks with noisy labels,'' \emph{Advances in neural information
  processing systems}, vol.~31, 2018.

\bibitem{Deng2021}
J.~Deng, C.~Gao, Q.~Feng, X.~Xu, and Z.~Chen, ``Adaptive generalized
  cross-entropy loss for sound event classification with noisy labels,'' in
  \emph{2021 IEEE Workshop on Applications of Signal Processing to Audio and
  Acoustics (WASPAA)}.\hskip 1em plus 0.5em minus 0.4em\relax IEEE, 2021, pp.
  256--260.

\bibitem{Heckelei2008}
T.~Heckelei, R.~C. Mittelhammer, and T.~Jansson, ``A bayesian alternative to
  generalized cross entropy solutions for underdetermined econometric models,''
  Tech. Rep., 2008.

\bibitem{Kazemdehdashti2018}
A.~Kazemdehdashti, M.~Mohammadi, and A.~R. Seifi, ``The generalized
  cross-entropy method in probabilistic optimal power flow,'' \emph{IEEE
  Transactions on Power Systems}, vol.~33, no.~5, pp. 5738--5748, 2018.

\bibitem{Kurian2021}
N.~C. Kurian, P.~S. Meshram, A.~Patil, S.~Patel, and A.~Sethi, ``Sample
  specific generalized cross entropy for robust histology image
  classification,'' in \emph{2021 IEEE 18th International Symposium on
  Biomedical Imaging (ISBI)}.\hskip 1em plus 0.5em minus 0.4em\relax IEEE,
  2021, pp. 1934--1938.

\bibitem{Jacobsen2018}
J.-H. Jacobsen, J.~Behrmann, R.~Zemel, and M.~Bethge, ``Excessive invariance
  causes adversarial vulnerability,'' \emph{Proceedings of the 7th
  International Conference on Learning Representations (ICLR), 2019}, Nov.
  2018.

\bibitem{Nar2019}
K.~Nar, O.~Ocal, S.~S. Sastry, and K.~Ramchandran, ``Cross-entropy loss and
  low-rank features have responsibility for adversarial examples,'' Jan. 2019.

\bibitem{Goodfellow2014}
I.~J. Goodfellow, J.~Shlens, and C.~Szegedy, ``Explaining and harnessing
  adversarial examples,'' 2014.

\bibitem{Madry2017}
A.~Madry, A.~Makelov, L.~Schmidt, D.~Tsipras, and A.~Vladu, ``Towards deep
  learning models resistant to adversarial attacks,'' 2017.

\bibitem{Carlini2017}
N.~Carlini and D.~Wagner, ``Towards evaluating the robustness of neural
  networks,'' in \emph{2017 IEEE Symposium on Security and Privacy (SP)}, 2017,
  pp. 39--57.

\bibitem{Schmidt2018}
\BIBentryALTinterwordspacing
L.~Schmidt, S.~Santurkar, D.~Tsipras, K.~Talwar, and A.~Madry, ``Adversarially
  robust generalization requires more data,'' in \emph{Advances in Neural
  Information Processing Systems}, S.~Bengio, H.~Wallach, H.~Larochelle,
  K.~Grauman, N.~Cesa-Bianchi, and R.~Garnett, Eds., vol.~31.\hskip 1em plus
  0.5em minus 0.4em\relax Curran Associates, Inc., 2018. [Online]. Available:
  \url{https://proceedings.neurips.cc/paper/2018/file/f708f064faaf32a43e4d3c784e6af9ea-Paper.pdf}
\BIBentrySTDinterwordspacing

\bibitem{Krizhevsky2009}
A.~Krizhevsky, ``Learning multiple layers of features from tiny images,'' 2009.

\bibitem{Zhang2019}
\BIBentryALTinterwordspacing
H.~Zhang, Y.~Yu, J.~Jiao, E.~Xing, L.~E. Ghaoui, and M.~Jordan, ``Theoretically
  principled trade-off between robustness and accuracy,'' in \emph{Proceedings
  of the 36th International Conference on Machine Learning}, ser. Proceedings
  of Machine Learning Research, K.~Chaudhuri and R.~Salakhutdinov, Eds.,
  vol.~97.\hskip 1em plus 0.5em minus 0.4em\relax PMLR, 09--15 Jun 2019, pp.
  7472--7482. [Online]. Available:
  \url{https://proceedings.mlr.press/v97/zhang19p.html}
\BIBentrySTDinterwordspacing

\bibitem{Tramer2017}
F.~Tramèr, A.~Kurakin, N.~Papernot, I.~Goodfellow, D.~Boneh, and P.~McDaniel,
  ``Ensemble adversarial training: Attacks and defenses,'' 2017.

\bibitem{Taghanaki2019}
S.~A. Taghanaki, K.~Abhishek, S.~Azizi, and G.~Hamarneh, ``A kernelized
  manifold mapping to diminish the effect of adversarial perturbations,'' in
  \emph{Proceedings of the IEEE/CVF Conference on Computer Vision and Pattern
  Recognition (CVPR)}, June 2019.

\bibitem{Zoran2020}
D.~Zoran, M.~Chrzanowski, P.-S. Huang, S.~Gowal, A.~Mott, and P.~Kohli,
  ``Towards robust image classification using sequential attention models,'' in
  \emph{Proceedings of the IEEE/CVF Conference on Computer Vision and Pattern
  Recognition (CVPR)}, June 2020.

\bibitem{Vaishnavi2020}
P.~Vaishnavi, T.~Cong, K.~Eykholt, A.~Prakash, and A.~Rahmati, ``Can attention
  masks improve adversarial robustness?'' in \emph{Engineering Dependable and
  Secure Machine Learning Systems}, O.~Shehory, E.~Farchi, and G.~Barash,
  Eds.\hskip 1em plus 0.5em minus 0.4em\relax Cham: Springer International
  Publishing, 2020, pp. 14--22.

\bibitem{Sokolic2017}
J.~Sokolić, R.~Giryes, G.~Sapiro, and M.~R.~D. Rodrigues, ``Robust large
  margin deep neural networks,'' \emph{IEEE Transactions on Signal Processing},
  vol.~65, no.~16, pp. 4265--4280, 2017.

\bibitem{Cisse2017}
\BIBentryALTinterwordspacing
M.~Cisse, P.~Bojanowski, E.~Grave, Y.~Dauphin, and N.~Usunier, ``Parseval
  networks: Improving robustness to adversarial examples,'' in
  \emph{Proceedings of the 34th International Conference on Machine Learning},
  ser. Proceedings of Machine Learning Research, D.~Precup and Y.~W. Teh, Eds.,
  vol.~70.\hskip 1em plus 0.5em minus 0.4em\relax PMLR, 06--11 Aug 2017, pp.
  854--863. [Online]. Available:
  \url{https://proceedings.mlr.press/v70/cisse17a.html}
\BIBentrySTDinterwordspacing

\bibitem{Kurakin2018}
A.~Kurakin, I.~Goodfellow, S.~Bengio, Y.~Dong, F.~Liao, M.~Liang, T.~Pang,
  J.~Zhu, X.~Hu, C.~Xie, J.~Wang, Z.~Zhang, Z.~Ren, A.~Yuille, S.~Huang,
  Y.~Zhao, Y.~Zhao, Z.~Han, J.~Long, Y.~Berdibekov, T.~Akiba, S.~Tokui, and
  M.~Abe, ``Adversarial attacks and defences competition,'' in \emph{The NIPS
  '17 Competition: Building Intelligent Systems}, S.~Escalera and M.~Weimer,
  Eds.\hskip 1em plus 0.5em minus 0.4em\relax Cham: Springer International
  Publishing, 2018, pp. 195--231.

\bibitem{Pang2019}
\BIBentryALTinterwordspacing
T.~Pang, K.~Xu, C.~Du, N.~Chen, and J.~Zhu, ``Improving adversarial robustness
  via promoting ensemble diversity,'' in \emph{Proceedings of the 36th
  International Conference on Machine Learning}, ser. Proceedings of Machine
  Learning Research, K.~Chaudhuri and R.~Salakhutdinov, Eds., vol.~97.\hskip
  1em plus 0.5em minus 0.4em\relax PMLR, 09--15 Jun 2019, pp. 4970--4979.
  [Online]. Available: \url{https://proceedings.mlr.press/v97/pang19a.html}
\BIBentrySTDinterwordspacing

\bibitem{Pang2019a}
T.~Pang, K.~Xu, Y.~Dong, C.~Du, N.~Chen, and J.~Zhu, ``Rethinking softmax
  cross-entropy loss for adversarial robustness,'' 2019.

\bibitem{Mustafa2019}
A.~Mustafa, S.~Khan, M.~Hayat, R.~Goecke, J.~Shen, and L.~Shao, ``Adversarial
  defense by restricting the hidden space of deep neural networks,'' in
  \emph{Proceedings of the IEEE/CVF International Conference on Computer Vision
  (ICCV)}, October 2019.

\bibitem{Pang2018}
\BIBentryALTinterwordspacing
T.~Pang, C.~Du, and J.~Zhu, ``Max-{M}ahalanobis linear discriminant analysis
  networks,'' in \emph{Proceedings of the 35th International Conference on
  Machine Learning}, ser. Proceedings of Machine Learning Research, J.~Dy and
  A.~Krause, Eds., vol.~80.\hskip 1em plus 0.5em minus 0.4em\relax PMLR, 10--15
  Jul 2018, pp. 4016--4025. [Online]. Available:
  \url{https://proceedings.mlr.press/v80/pang18a.html}
\BIBentrySTDinterwordspacing

\bibitem{Amid2019}
E.~Amid, M.~K.~K. Warmuth, R.~Anil, and T.~Koren, ``Robust bi-tempered logistic
  loss based on bregman divergences,'' in \emph{Advances in Neural Information
  Processing Systems}, H.~Wallach, H.~Larochelle, A.~Beygelzimer,
  F.~d\textquotesingle Alch\'{e}-Buc, E.~Fox, and R.~Garnett, Eds.,
  vol.~32.\hskip 1em plus 0.5em minus 0.4em\relax Curran Associates, Inc.,
  2019.

\bibitem{Lin2017}
T.-Y. Lin, P.~Goyal, R.~Girshick, K.~He, and P.~Dollar, ``Focal loss for dense
  object detection,'' in \emph{Proceedings of the IEEE International Conference
  on Computer Vision (ICCV)}, Oct 2017.

\bibitem{Pezeshki2021}
\BIBentryALTinterwordspacing
M.~Pezeshki, O.~Kaba, Y.~Bengio, A.~C. Courville, D.~Precup, and G.~Lajoie,
  ``Gradient starvation: A learning proclivity in neural networks,'' in
  \emph{Advances in Neural Information Processing Systems}, M.~Ranzato,
  A.~Beygelzimer, Y.~Dauphin, P.~Liang, and J.~W. Vaughan, Eds., vol.~34.\hskip
  1em plus 0.5em minus 0.4em\relax Curran Associates, Inc., 2021, pp.
  1256--1272. [Online]. Available:
  \url{https://proceedings.neurips.cc/paper/2021/file/0987b8b338d6c90bbedd8631bc499221-Paper.pdf}
\BIBentrySTDinterwordspacing

\bibitem{Xiao2017}
H.~Xiao, K.~Rasul, and R.~Vollgraf, ``Fashion-mnist: a novel image dataset for
  benchmarking machine learning algorithms,'' \emph{arXiv preprint
  arXiv:1708.07747}, 2017.

\bibitem{Darlow2018}
L.~N. Darlow, E.~J. Crowley, A.~Antoniou, and A.~J. Storkey, ``Cinic-10 is not
  imagenet or cifar-10,'' 2018.

\bibitem{Xiao2015}
T.~Xiao, T.~Xia, Y.~Yang, C.~Huang, and X.~Wang, ``Learning from massive noisy
  labeled data for image classification,'' in \emph{Proceedings of the IEEE
  conference on computer vision and pattern recognition}, 2015, pp. 2691--2699.

\bibitem{Everingham}
M.~Everingham, L.~Van~Gool, C.~K.~I. Williams, J.~Winn, and A.~Zisserman, ``The
  {PASCAL} {V}isual {O}bject {C}lasses {C}hallenge 2007 {(VOC2007)}
  {R}esults,''
  http://www.pascal-network.org/challenges/VOC/voc2007/workshop/index.html.

\bibitem{Keskar2016}
N.~S. Keskar, D.~Mudigere, J.~Nocedal, M.~Smelyanskiy, and P.~T.~P. Tang, ``On
  large-batch training for deep learning: Generalization gap and sharp
  minima,'' 2016.

\bibitem{Prabhu2019}
V.~U. Prabhu, D.~A. Yap, J.~Xu, and J.~Whaley, ``Understanding adversarial
  robustness through loss landscape geometries,'' 2019.

\bibitem{Li2018}
\BIBentryALTinterwordspacing
H.~Li, Z.~Xu, G.~Taylor, C.~Studer, and T.~Goldstein, ``Visualizing the loss
  landscape of neural nets,'' in \emph{Advances in Neural Information
  Processing Systems}, S.~Bengio, H.~Wallach, H.~Larochelle, K.~Grauman,
  N.~Cesa-Bianchi, and R.~Garnett, Eds., vol.~31.\hskip 1em plus 0.5em minus
  0.4em\relax Curran Associates, Inc., 2018. [Online]. Available:
  \url{https://proceedings.neurips.cc/paper/2018/file/a41b3bb3e6b050b6c9067c67f663b915-Paper.pdf}
\BIBentrySTDinterwordspacing

\bibitem{He2016a}
K.~He, X.~Zhang, S.~Ren, and J.~Sun, ``Identity mappings in deep residual
  networks,'' in \emph{Computer Vision -- ECCV 2016}, B.~Leibe, J.~Matas,
  N.~Sebe, and M.~Welling, Eds.\hskip 1em plus 0.5em minus 0.4em\relax Cham:
  Springer International Publishing, 2016, pp. 630--645.

\bibitem{Wen2016a}
Y.~Wen, K.~Zhang, Z.~Li, and Y.~Qiao, ``A discriminative feature learning
  approach for deep face recognition,'' in \emph{Computer Vision -- ECCV 2016},
  B.~Leibe, J.~Matas, N.~Sebe, and M.~Welling, Eds.\hskip 1em plus 0.5em minus
  0.4em\relax Cham: Springer International Publishing, 2016, pp. 499--515.

\bibitem{Wan2018}
W.~Wan, Y.~Zhong, T.~Li, and J.~Chen, ``Rethinking feature distribution for
  loss functions in image classification,'' in \emph{Proceedings of the IEEE
  Conference on Computer Vision and Pattern Recognition (CVPR)}, June 2018.

\bibitem{He2009}
H.~He and E.~A. Garcia, ``Learning from imbalanced data,'' \emph{IEEE
  Transactions on Knowledge and Data Engineering}, vol.~21, no.~9, pp.
  1263--1284, 2009.

\bibitem{Elsayed2018}
\BIBentryALTinterwordspacing
G.~Elsayed, D.~Krishnan, H.~Mobahi, K.~Regan, and S.~Bengio, ``Large margin
  deep networks for classification,'' in \emph{Advances in Neural Information
  Processing Systems}, S.~Bengio, H.~Wallach, H.~Larochelle, K.~Grauman,
  N.~Cesa-Bianchi, and R.~Garnett, Eds., vol.~31.\hskip 1em plus 0.5em minus
  0.4em\relax Curran Associates, Inc., 2018. [Online]. Available:
  \url{https://proceedings.neurips.cc/paper/2018/file/42998cf32d552343bc8e460416382dca-Paper.pdf}
\BIBentrySTDinterwordspacing

\bibitem{Wang2017}
F.~Wang, M.~Jiang, C.~Qian, S.~Yang, C.~Li, H.~Zhang, X.~Wang, and X.~Tang,
  ``Residual attention network for image classification,'' in \emph{Proceedings
  of the IEEE Conference on Computer Vision and Pattern Recognition (CVPR)},
  July 2017.

\bibitem{Vaswani2017}
\BIBentryALTinterwordspacing
A.~Vaswani, N.~Shazeer, N.~Parmar, J.~Uszkoreit, L.~Jones, A.~N. Gomez, L.~u.
  Kaiser, and I.~Polosukhin, ``Attention is all you need,'' in \emph{Advances
  in Neural Information Processing Systems}, I.~Guyon, U.~V. Luxburg,
  S.~Bengio, H.~Wallach, R.~Fergus, S.~Vishwanathan, and R.~Garnett, Eds.,
  vol.~30.\hskip 1em plus 0.5em minus 0.4em\relax Curran Associates, Inc.,
  2017. [Online]. Available:
  \url{https://proceedings.neurips.cc/paper/2017/file/3f5ee243547dee91fbd053c1c4a845aa-Paper.pdf}
\BIBentrySTDinterwordspacing

\bibitem{Zhou2018}
Z.~Zhou, M.~M. Rahman~Siddiquee, N.~Tajbakhsh, and J.~Liang, ``Unet++: A nested
  u-net architecture for medical image segmentation,'' in \emph{Deep Learning
  in Medical Image Analysis and Multimodal Learning for Clinical Decision
  Support}, D.~Stoyanov, Z.~Taylor, G.~Carneiro, T.~Syeda-Mahmood, A.~Martel,
  L.~Maier-Hein, J.~M.~R. Tavares, A.~Bradley, J.~P. Papa, V.~Belagiannis,
  J.~C. Nascimento, Z.~Lu, S.~Conjeti, M.~Moradi, H.~Greenspan, and
  A.~Madabhushi, Eds.\hskip 1em plus 0.5em minus 0.4em\relax Cham: Springer
  International Publishing, 2018, pp. 3--11.

\bibitem{Chen2020}
\BIBentryALTinterwordspacing
J.~Chen and T.~Bai, ``Saanet: Spatial adaptive alignment network for object
  detection in automatic driving,'' \emph{Image and Vision Computing}, vol.~94,
  p. 103873, 2020. [Online]. Available:
  \url{https://www.sciencedirect.com/science/article/pii/S0262885620300056}
\BIBentrySTDinterwordspacing

\bibitem{Lu2020}
W.~Lu, J.~Li, J.~Wang, and L.~Qin, ``A {CNN}-{BiLSTM}-{AM} method for stock
  price prediction,'' \emph{Neural Computing and Applications}, vol.~33,
  no.~10, pp. 4741--4753, nov 2020.

\bibitem{Goodfellow2014a}
\BIBentryALTinterwordspacing
I.~Goodfellow, J.~Pouget-Abadie, M.~Mirza, B.~Xu, D.~Warde-Farley, S.~Ozair,
  A.~Courville, and Y.~Bengio, ``Generative adversarial nets,'' in
  \emph{Advances in Neural Information Processing Systems}, Z.~Ghahramani,
  M.~Welling, C.~Cortes, N.~Lawrence, and K.~Weinberger, Eds., vol.~27.\hskip
  1em plus 0.5em minus 0.4em\relax Curran Associates, Inc., 2014. [Online].
  Available:
  \url{https://proceedings.neurips.cc/paper/2014/file/5ca3e9b122f61f8f06494c97b1afccf3-Paper.pdf}
\BIBentrySTDinterwordspacing

\bibitem{Srivastava2014}
N.~Srivastava, G.~Hinton, A.~Krizhevsky, I.~Sutskever, and R.~Salakhutdinov,
  ``Dropout: A simple way to prevent neural networks from overfitting,''
  \emph{J. Mach. Learn. Res.}, vol.~15, no.~1, p. 1929–1958, jan 2014.

\bibitem{He2016}
K.~He, X.~Zhang, S.~Ren, and J.~Sun, ``Deep residual learning for image
  recognition,'' in \emph{Proceedings of the IEEE Conference on Computer Vision
  and Pattern Recognition (CVPR)}, June 2016.

\bibitem{Wen2016}
Y.~Wen, K.~Zhang, Z.~Li, and Y.~Qiao, ``A discriminative feature learning
  approach for deep face recognition,'' in \emph{Computer Vision -- ECCV 2016},
  B.~Leibe, J.~Matas, N.~Sebe, and M.~Welling, Eds.\hskip 1em plus 0.5em minus
  0.4em\relax Cham: Springer International Publishing, 2016, pp. 499--515.

\bibitem{Zhao2020}
H.~Zhao, J.~Jia, and V.~Koltun, ``Exploring self-attention for image
  recognition,'' in \emph{Proceedings of the IEEE/CVF Conference on Computer
  Vision and Pattern Recognition (CVPR)}, June 2020.

\bibitem{Matyasko2017}
A.~Matyasko and L.-P. Chau, ``Margin maximization for robust classification
  using deep learning,'' in \emph{2017 International Joint Conference on Neural
  Networks (IJCNN)}.\hskip 1em plus 0.5em minus 0.4em\relax IEEE, 2017, pp.
  300--307.

\bibitem{Weinberger2009}
K.~Q. Weinberger and L.~K. Saul, ``Distance metric learning for large margin
  nearest neighbor classification.'' \emph{Journal of machine learning
  research}, vol.~10, no.~2, 2009.

\bibitem{Sun2016}
S.~Sun, W.~Chen, L.~Wang, X.~Liu, and T.-Y. Liu, ``On the depth of deep neural
  networks: A theoretical view,'' \emph{Proceedings of the {AAAI} Conference on
  Artificial Intelligence}, vol.~30, no.~1, mar 2016.

\bibitem{Bergeler2015}
M.~Bergeler, C.~Herrmann, and M.~Reiher, ``Mode-tracking based stationary-point
  optimization,'' \emph{Journal of Computational Chemistry}, vol.~36, no.~19,
  pp. 1429--1438, jun 2015.

\bibitem{Bergeler2015a}
\BIBentryALTinterwordspacing
------, ``Mode-tracking based stationary-point optimization,'' \emph{Journal of
  Computational Chemistry}, vol.~36, no.~19, pp. 1429--1438, 2015. [Online].
  Available: \url{https://onlinelibrary.wiley.com/doi/abs/10.1002/jcc.23958}
\BIBentrySTDinterwordspacing

\bibitem{VandeVijver2020}
\BIBentryALTinterwordspacing
R.~{Van de Vijver} and J.~Zádor, ``Kinbot: Automated stationary point search
  on potential energy surfaces,'' \emph{Computer Physics Communications}, vol.
  248, p. 106947, 2020. [Online]. Available:
  \url{https://www.sciencedirect.com/science/article/pii/S0010465519302978}
\BIBentrySTDinterwordspacing

\bibitem{Park2019}
J.~Park, S.~Park, D.-H. Kim, and S.-O. Park, ``Leakage mitigation in heterodyne
  fmcw radar for small drone detection with stationary point concentration
  technique,'' \emph{IEEE Transactions on Microwave Theory and Techniques},
  vol.~67, no.~3, pp. 1221--1232, 2019.

\end{thebibliography}
\end{document}